\documentclass[letterpaper, 10 pt, conference]{ieeeconf}

\usepackage[table]{xcolor} 
\usepackage{graphicx}
\usepackage{float}
\usepackage{cite}
\usepackage{booktabs}
\usepackage{amsmath}
\usepackage[ruled,vlined]{algorithm2e}

\usepackage{mdframed}
\usepackage{tcolorbox}
\tcbuselibrary{listings, breakable}
\usepackage{url}
\usepackage{amsmath, amssymb, graphicx}
\usepackage{makecell}
\usepackage{siunitx} 
\usepackage{etoolbox}

\makeatletter
\patchcmd{\@algocf@makecaption@plain}
  {\hbox}
  {\colorbox{gray!20}{\hbox}}
  {}{}
\makeatother

\usepackage[colorlinks=true, linkcolor=black, citecolor=black, urlcolor=black]{hyperref}
\hypersetup{
    pdfborder={0 0 0}, 
    breaklinks=true
}
\IEEEoverridecommandlockouts                      
\title{\LARGE \bf
CRADMap: Applied Distributed Volumetric Mapping with 5G-Connected Multi-Robots and 4D Radar Perception
}

\author{Maaz Qureshi$^{1}$*, Alexander Werner$^{1}$, Zhenan Liu$^{1}$, Amir Khajepour$^{1}$, George Shaker$^{1}$, and William Melek$^{1}$
\thanks{Work is supported by Rogers Communication Canada Inc and Mitacs.}
\thanks{$^{1}$All authors are with the Faculty of Mechanical and Mechatronics Engineering,
        University of Waterloo (UW), 200 University Ave W, Waterloo, ON N2L3G1, Canada.   *Corresponding-Author Email: {\tt\small m23qures@uwaterloo.ca}}%
}

\begin{document}

\maketitle
\thispagestyle{empty}
\pagestyle{empty}

\begin{abstract}

Sparse and feature SLAM methods provide robust camera pose estimation. However, they often fail to capture the level of detail required for inspection and scene awareness tasks. Conversely, dense SLAM approaches generate richer scene reconstructions but impose a prohibitive computational load to create 3D maps. We present a novel distributed volumetric mapping framework designated as CRADMap that addresses these issues by extending the state-of-the-art (SOTA) ORBSLAM3 system with the COVINS on the backend for global optimization. Our pipeline for volumetric reconstruction fuses dense keyframes at a centralized server via 5G connectivity, aggregating geometry, and occupancy information from multiple autonomous mobile robots (AMRs) without overtaxing onboard resources. This enables each AMR to independently perform mapping while the backend constructs high-fidelity real-time 3D maps. To operate Beyond the Visible (BtV) and overcome the limitations of standard visual sensors, we automated a standalone 4D mmWave radar module that functions independently without sensor fusion with SLAM. The BtV system enables the detection and mapping of occluded metallic objects in cluttered environments, enhancing situational awareness in inspection scenarios. Experimental validation in Section~\ref{sec:IV} demonstrates the effectiveness of our framework.
\href{https://youtu.be/eTLxCY2rRMA}{\textit{ Video Attachment: https://youtu.be/eTLxCY2rRMA}}

\end{abstract}

\section{INTRODUCTION}\label{sec:I}

The rapid development of robot autonomy has opened up new opportunities to automate routine tasks and improve safety in hazardous environments. With reliable simultaneous localization and mapping (SLAM) systems robust in pose estimation and map generation, there is a growing gap in multi-robot distributed SLAM to build detailed 3D maps using low-cost autonomous mobile robots (AMRs), with limited processing capabilities, for real-time and efficient mapping. Such systems are essential for applications such as  navigation, path planning, and inspection. Mapping large complex indoor environments presents several unresolved problems. First, coordinating data from multiple robots to generate a dense, volumetric map requires significant computational power. Second, handling overlapping sensor data from multiple AMRs while maintaining distributed maps accuracy remains a critical issue. Third, although high-speed data transmission from WiFi offers great potential, managing high bandwidth with low latency for real-time performance is not trivial.

\begin{figure}[htbp]
    \centerline{\includegraphics[width=0.48\textwidth]{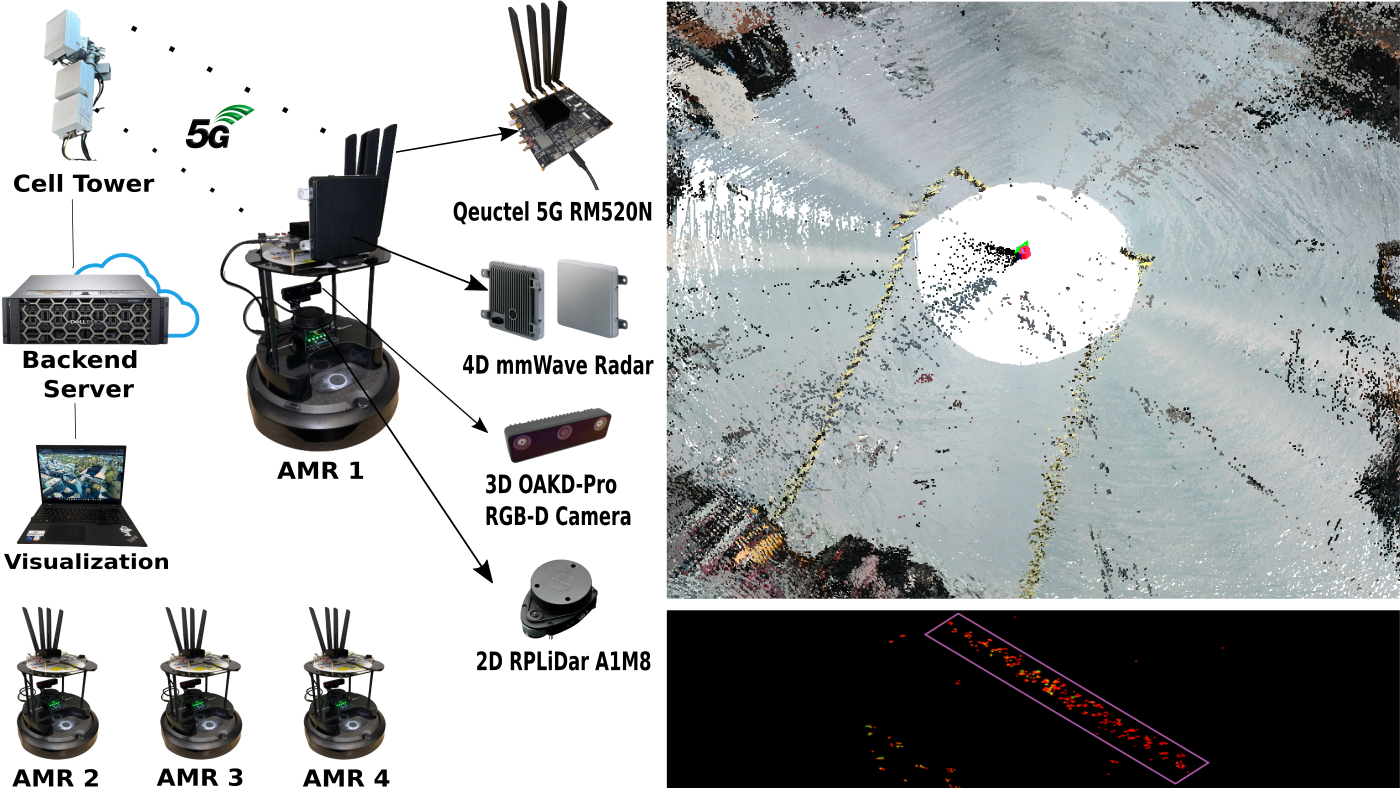}}
    \caption{Overview of our approach. The left section shows the data transmission cycle for map visualization. The center highlights an AMR \cite{turtlebot4} with its key integrated components. The right section presents a 360-degree top-view CRADMap of the UW RoboHub lab, while the bottom right shows the radar point cloud map detecting an obscured vent pipe.}
    \label{fig1}
    \vspace*{-0.3cm}
\end{figure}

Industries such as manufacturing, warehouse management, and asset or infrastructure inspection, which rely on autonomous robotics in damage monitoring, digital inspections, and material handling, require the deployment of reliable solutions to enable the efficient completion of the aforementioned tasks. These environments are large and often cluttered with obstacles like machinery, walls, and structural elements, which can obstruct the view of visual sensors. This limitation affects tasks that require precise navigation and detailed mapping with a team of AMRs. As industrial applications become more demanding, there is a growing need for systems that can effectively manage these complexities and deliver reliable real-time mapping.

\begin{figure*}[]
\centerline{\includegraphics[width=1.0\textwidth]{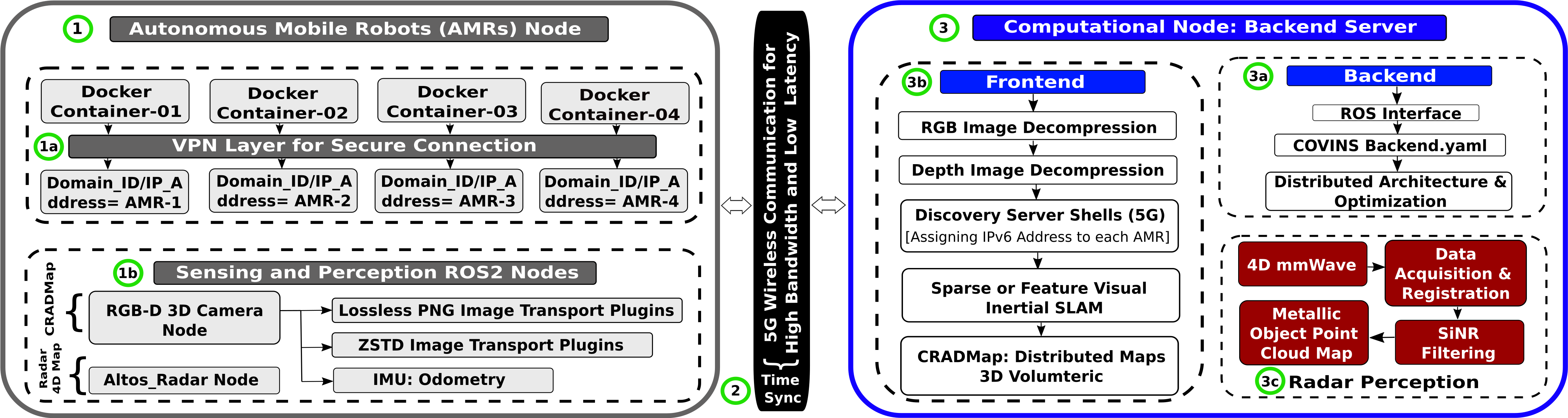}}
    \caption{Comprehensive pipeline flow of the proposed methodology in three modules. The gray module illustrates the structure of the AMRs nodes. The blue module represents the frontend and backend stages of the distributed framework for multi-robot volumetric maps and radar point cloud generation, which are managed on a central server to offload computational processing. The black layer in the middle highlights the real-time data transmission stage.}
    \label{fig2}
     \vspace*{-0.5cm}
\end{figure*}

In response to these challenges, we develop CRADMap, a novel distributed volumetric mapping architecture that scales sparse, feature-based SLAM on the frontend with volumetric keyframe fusion on the backend cloud server, an overview shown in Figure. \ref{fig1}. In our approach, we enhance keyframes so that they not only store the camera pose (i.e., its position and orientation) and visual features but also include a complete 3D dense point cloud. This enriched data is transmitted over 5G to achieve high bandwidth with low latency to the backend server, which fuses keyframes from view, integrates dense point clouds, and reconstructs the scene to generate globally consistent, high-fidelity 3D maps in real-time. Additionally, since visual nodes miss objects hidden by obstacles, we automated an independent 4D mmWave radar that creates its point map to detect occluded or hidden metallic objects.

The main contributions of this work are as follows:
\begin{itemize}
  \item  Mapping framework (CRADMap) that enhances Visual-SLAM with enriched keyframes containing dense point clouds to generate volumetric maps in real-time.
  \item A multi-robot distributed architecture that leverages a high-bandwidth i.e. 5G network to offload intensive processing to a centralized COVINS backend, ensuring real-time, globally optimized 3D reconstructions.
  \item Integration of an independent 4D mmWave radar that creates a standalone point cloud map, supplementing CRADMap without fusion with RGB-D for Beyond the Visible (BtV) detection of hidden structures.
 \item Evaluation on live indoor experiments from multiple AMRs to demonstrate improvements and efficiency.
\end{itemize}

The paper is organized as follows. Section \ref{sec:II} discusses related work, and Section \ref{sec:III Methods} describes the overall system methods. Experiments are presented in Section \ref{sec:IV}.  Lastly, Section \ref{sec:V} presents the conclusion with future work.

\section{RELATED WORK} \label{sec:II}
\subsection {Volumetric mapping}
Early RGB-D SLAM efforts by Henry et al. \cite{Henry2014} and Endres et al. \cite{article}  achieved dense reconstructions yet often relied on offline optimization and high-end GPUs, making them impractical for real-time volumetric mapping. KinectFusion \cite{6162880} introduced TSDF-based mapping for smaller indoor spaces, leveraging powerful hardware to sustain real-time performance. Later works such as voxel hashing by Nießner et al. \cite{Niener2013Realtime3R} scaled KinectFusion to larger scenes, while Oleynikova et al. \cite{8202315}  (Voxblox) and Millane et al. \cite{8903279} (Voxgraph) refined incremental fusion and drift correction mainly in single-AMR or offline settings. Additionally, an octree-based approach by Hou et al. \cite{hou2023octree} processed RGB-D video for real-time 3D indoor mapping, but it remained limited to a single-AMR pipeline without addressing the complexity of multi-robot collaboration. As a result, large-scale 3D volumetric maps on low-cost AMRs remain a key challenge, especially under live conditions.

\subsection{Multi-Robots SLAM Techniques and 5G Communication}
Modern distributed and decentralized SLAM solutions such as Kimera-Multi \cite{9686955} and Swarm-SLAM \cite{10321649} extend multi-robot mappings frameworks advance multi-agent coordination but demand significant GPU resources on limited bandwidth networks. Meanwhile, CCM-SLAM \cite{schmuck2019ccm} and COVINS \cite{schmuck2021covins} reduce global drift by merging keyframes on a centralized backend server, primarily handling sparse data. Our approach extends ORBSLAM3 \cite{campos2021orb} with COVINS \cite{schmuck2021covins}, showing that centralized global optimization enhances keyframe accuracy for multi-robot in distributed architecture instead of map merging. To handle dense volumetric data from multiple robots in real-time, we utilize 5G communication, as highlighted by recent work \cite{komatsu2022leveraging} demonstrating how high throughput and low latency enable offloading of large sensor streams. By transmitting dense keyframes over 5G, our system alleviates each AMR computing load, allowing real-time distributed volumetric reconstruction.

\subsection{4D mmWave Perception Technology}
Although radar-based SLAM \cite{10175555}, \cite{10160670} can improve navigation in harsh visibility i.e. fog, snow, and low lighting, it introduces complex sensor fusion, and potential noise not present in indoor environments. Motivated from Doojin Lee et al \cite{lee2020imaging} and Wallabot \cite{walabot} illustrate the radar unique capability for detecting metallic or hidden objects behind walls. In our framework, radar operates separately from the CRADMap pipeline, focusing on detecting obscured objects that are not visible by camera or lidar. This modular design maintains a streamlined volumetric map referred to here as CRADMap and in addition, delivers a standalone radar point cloud map for detecting metal items that are obscured behind the structure and walls.

\begin{table*}[]
\caption{5G NR Frequency Band Parameters \cite{sqimway_nr_bands} for Time Division Duplexing (TDD) Communication}
\centering
\label{table:1}
\begin{tabular}{cccccccc}
\hline
\rowcolor{gray!20}
\textbf{Band} & \textbf{Name} & \textbf{Mode} & \textbf{$\Delta$FRaster (kHz)} & \textbf{Low (MHz)} & \textbf{Middle (MHz)} & \textbf{High (MHz)} & \textbf{DL/UL Bandwidth (MHz)} \\ \hline
78            & TD 3500       & TDD           & 15, 30                         & 3300                    & 3550                     & 3800                     & 500                            \\ \hline
\end{tabular}
\end{table*}

\section{METHODS} \label{sec:III Methods}
Our proposed system, CRADMap, is a distributed multi-robot framework for applied volumetric mapping that uses ORBSLAM3 on front-ends, a centralized COVINS on back-end for distributed architecture, and a dedicated Altos 4D mmWave radar \cite{altosradar2023} to generate additional point cloud map of obscured metallic object(s). Fig. \ref{fig2} illustrates the detailed process blocks involved. The system leverages high-bandwidth 5G connectivity ($\approx$24 ms end-to-end latency, up to 110 MBps) to enable four AMRs (autonomous mobile robots) as shown in Fig. \ref{fig1}. Each integrated with quectel 5G RM520N \cite{quectel_rm520n} to stream data in real-time with OAK-D pro \cite{oakdpro2024} 3D RBG-D camera, with inherent scalability to $N$ AMRs. In the following, we describe our research pipeline supported by algorithms to illustrate the flow of implementation.

\subsection{Front-End: ORBSLAM3 with Dense Keyframe Generation} \label{sec:IIIA}
Each AMR processes an RGB-D data stream $(I_t, D_t)$ at time $t$ using ORB-SLAM3 to extract features and estimate camera poses. Our method enriches standard ORB-SLAM3 keyframes by generating dense point clouds, as outlined in Algorithm~\ref{alg:keyframe}, to facilitate robust volumetric mapping.

\subsection*{i. Feature Extraction and Pose Estimation}
\subsubsection*{a) ORB Feature Extraction}
For each frame $I_t$, ORBSLAM3 extracts 2D keypoints $\{f_{ij}\}$ (with $i$ indexing keyframes and $j$ indexing keypoints).

\subsubsection*{b) Depth Association}
For each detected keypoint at pixel $u_{ij}$, the depth $d_{ij}$ is retrieved from the depth image $D_t$. The corresponding 3D point $x_{ij}$ is computed via:
\begin{equation}
    x_{ij} = \Pi^{-1}(u_{ij}, d_{ij}) = d_{ij} \cdot K^{-1} \begin{bmatrix} u_{ij} \\ 1 \end{bmatrix}
\label{eq:1}
\end{equation}
where $\Pi^{-1}$ is the back-projection operator using camera intrinsics ($K$).

\subsubsection*{c) Pose Estimation}
The camera pose $T_t \in SE(3)$ is obtained by minimizing the reprojection error:
\begin{equation}
T_t = \arg\min_{T \in SE(3)} \sum_{i,j} \| u_{ij} - \pi(T \cdot x_{ij}) \|^2
\label{eq:2}
\end{equation}
\vspace{-1pt}
Here, $\pi(\cdot)$ projects a 3D point onto the image plane, ensuring alignment between observed key points and reprojected 3D points. SE(3) is a special Euclidean group in three dimensions.

\subsection*{ii. Dense Keyframe Generation}
\subsubsection*{d) Dense Point Cloud Construction}
When keyframe criteria (e.g., significant camera motion or scene change) are met, a dense point cloud is generated from the entire image:
\begin{equation}
P_i = \{x \mid x = \Pi^{-1}(u, D_i[u]), \, u \in \Omega(I_i)\}
\label{eq:3}
\end{equation}
where $\Omega(I_i)$ is the set of all pixel coordinates in image $I_i$.

\subsubsection*{e) Outlier Removal}
To improve robustness, we filter out noisy points by comparing each point $x$ to the local neighborhood statistics:
\begin{equation}
\|x - \mu_N(x)\| \leq \lambda \sigma_N(x)
\label{eq:4}
\end{equation}
where $\mu_N(x)$ and $\sigma_N(x)$ are the mean and standard deviation of neighboring points, and $\lambda$ is a threshold. The cleaned point cloud is denoted as $P_i^{\text{clean}}$.

\subsubsection*{f) Keyframe Packaging}
The keyframe is then stored as:
\begin{equation}
K_i = \{T_i, \{f_{ij}\}, I_i, D_i, P_i^{\text{clean}}\}
\label{eq:5}
\end{equation}
which includes the estimated pose $T_i$, features, RGB-D images, and the cleaned dense point cloud.
\subsection*{iii. Image Plugins and 5G Bandwidth}
For real-time streaming we use lossless PNG compression for RGB (15 FPS) and ZSTD for depth (10 FPS) ROS2 camera topics using image transport plugins \cite{ros_perception_image_transport_plugins}. Compressed images are sent to the backend server for pose optimization and distributed map handling in COVINS. This integration ensures the creation of accurate dense volumetric maps. Each AMR is assigned a unique IPv6 address via the 5G Wwan, enabling real-time streaming and SLAM processing. We operate in the NR FR1 channel details presented in Table \ref{table:1} which supports high data rates required for RGB-D. 5G ensures fast IPv6-based discovery and low latency. 

\subsubsection*{g) Bandwidth Calculation}
each-AMR bandwidth ($B_r$)
\begin{equation}
B_r = f \times (PNG_{\text{comp}}^{\text{RGB}} + ZSTD_{\text{comp}}^{\text{depth}})
\label{eq:6}
\end{equation}

Total bandwidth for four AMRs:
\vspace{-3pt}
\begin{equation}
B_{\text{total}} = \sum_{r=1}^{4} B_r \leq 110 \text{ Mbps}
\label{eq:7}
\end{equation}

\vspace{-5pt}

\begin{algorithm}[htbp]
\caption{\colorbox{gray!20}{\parbox{\dimexpr\linewidth-13\fboxsep}{
Dense Keyframe Generation}}}
\label{alg:keyframe}
\KwData{RGB-D stream $(I_t, D_t)$ for each frame $t$}
\KwResult{Stream of keyframes $\{K_i\}$}

\For{each frame $t$}{
    \textbf{Feature Extraction \& Depth Association}:\;
    Detect ORB keypoints in $I_t$\;
    \For{each detected keypoint $u_{ij}$}{
        Retrieve depth $d_{ij}$ from $D_t$\;
        Compute 3D point $x_{ij}$ (Eq.~\ref{eq:1})\;
    }
    
    \textbf{Pose Estimation}:\;
    Estimate $T_t$ by minimizing error (Eq.~\ref{eq:2})\;
    
    \textbf{Keyframe Decision \& Generation}\;
    \If{keyframe criteria are met}{
        Generate dense $P_i$ from $I_t$ and $D_t$ (Eq.~\ref{eq:3})\;
        Filter $P_i$ to obtain $P_i^{\text{clean}}$ (Eq.~\ref{eq:4})\;
        Package keyframe $K_i$ (Eq.~\ref{eq:5})\;
        Compress $I_t$ (PNG) and $D_t$ (zstd) (Eqs.~\ref{eq:6}, \ref{eq:7})\;
        Transmit $K_i$ over the 5G network\;
    }
}
\Return $\{K_i\}$\;
\end{algorithm}

\subsection{Back-End: COVINS for  Optimizations} \label{sec:IIIB}

The COVINS back-end receives keyframes from all AMRs, stores them in a global atlas, and performs loop closure and pose graph optimization to refine camera poses and reduce drift. By enforcing global consistency, COVINS improves both Absolute Trajectory Error (ATE) and Root Mean Square Error (RMSE). Unlike collaborative map merging, it maintains individual RGB-D maps from each AMR in a distributed manner, as detailed in Algorithm~\ref{alg:pose_optimization}.\label{IIIB}
\vspace{-5pt}
\subsection*{i. Keyframes Integration}a) Keyframes ($K_i$) are stored in the atlas as:
\begin{equation}
K_i = \{T_i, P_i^{\text{clean}}, \{f_{ij}\}\}
\label{eq:8}
\end{equation}
\vspace{-20pt}
\subsection*{ii. Loop Closure and Global Pose Optimization}
\subsubsection*{b) Loop Closure Constraint}
 For a new keyframe $K_n$, candidate keyframes $K_j$ are identified via a Bag-of-Words approach. A loop closure constraint is computed as:
\begin{equation}
C(T_n, T_j) = T_n^{-1} T_j
\label{eq:9}
\end{equation}
\vspace{-15pt}
\subsubsection*{c) Global Pose Optimization}
The global pose graph is refined by minimizing a cost function that combines odometry and loop closure constraints:
\begin{multline}
\{T_i^*\} = \arg\min_{\{T_i\}} \Biggl( \sum_{(i,k) \in E_{\text{odom}}} \|\Delta(T_i, T_k)\|_{\Sigma_{\text{odom}}}^2 \\
+ \sum_{(i,k) \in E_{\text{loop}}} \|C(T_i, T_k)\|_{\Sigma_{\text{loop}}}^2 \Biggr)
\label{eq:10}
\end{multline}
Here, $\Delta(T_i, T_k)$ represents the relative odometry error, and $\Sigma_{\text{odom}}$ and $\Sigma_{\text{loop}}$ are covariance matrices for odometry and loop closure constraints, respectively.

\begin{algorithm}[htbp]
\caption{\colorbox{gray!20}{\parbox{\dimexpr\linewidth-13\fboxsep}{Pose Optimization and Loop Closure}}}
\label{alg:pose_optimization}
\KwData{Set of keyframes $\{K_i\}$ from all AMRs}
\KwResult{Refined global pose graph $\{T_i^*\}$}

\textbf{Keyframe Integration}:\;
\For{each incoming keyframe $K_i$}{
    Store $K_i$ in the global atlas\;
}

\textbf{Loop Closure Detection}:\;
\For{each new keyframe $K_n$}{
    Identify candidate keyframes using Bag-of-Words\;
    \For{each candidate $K_j$}{
        Compute loop closure constraint (Eq.~\ref{eq:9})\;
    }
}

\textbf{Global Optimization}:\;
Optimize global pose graph by minimizing combined cost function (Eq.~\ref{eq:10})\;

\textbf{Update \& Broadcast}:\;
Refined poses $\{T_i^*\}$ to all AMRs\;
Update global atlas\;

\Return $\{T_i^*\}$\;
\end{algorithm}

\subsection{Distributed Volumetric Mapping} \label{sec:IIIC}
In CRADMap, each AMR constructs its individual volumetric map from dense keyframes, transformed into the global coordinate frame using refined poses from COVINS. The distributed framework operates as follows: each AMR independently processes its RGB-D stream, generates dense keyframes from frontend (\ref{sec:IIIA}), and builds a local map without map merging in a distributed manner (\ref{IIIB}). This pipeline ensures scalability by eliminating the need for computationally expensive map merging, and robustness against individual AMR failures. Each AMR $r$ generates a local map $M_r$ by transforming the dense point clouds from each keyframe into the global coordinate frame using the refined pose:
\begin{equation}
X = T_i^* \cdot x,  \forall x \in P_i
\end{equation}
Then, the local map is formed as the union of all transformed points:
\vspace{-5pt}
\begin{equation}
M_r (CRADMap) = \bigcup_{i \in K_r} \{X\}
\label{eq:11}
\end{equation}

\subsection*{4D Radar-Metallic Object Detection} \label{sec:IIID}
The 4D mmWave radar facilitates the detection of occluded metallic structures and autonomously triggers Algorithm~\ref{algo:three}. By weighting point cloud registration with SNR data and filtering non-metallic returns, the system enhances metallic detection for mapping hidden utilities, as described and evaluated in Section~\ref{IVD}.

\subsubsection*{i. Radar Data Acquisition and SNR Filtering}
At time $t$, the radar produces a point cloud $R_t = \{r_k\}$. Points are filtered based on SNR (Signal-to-Noise Ratio):
\begin{equation}
R_t^{\text{metal}} = \{r_k \in R_t \mid s_k \geq s_{\text{th}}\}
\label{eq:13}
\end{equation}

\subsubsection*{ii. Global Transformation and Noise Removal}
Filtered radar points are transformed to the global frame:
\begin{equation}
Z = T_i^* \cdot r_k \cdot 
\begin{bmatrix}
\cos\theta_k \cos\phi_k \\
\sin\theta_k \cos\phi_k \\
\sin\phi_k
\end{bmatrix}
\label{eq:14}
\end{equation}

A statistical outlier removal filter is then applied to $Z$:
\begin{equation}
M_{\text{radar}} = \{Z \mid Z \text{ passes noise filtering}\}
\label{eq:15}
\end{equation}

\begin{algorithm}[htbp]
\caption{\colorbox{gray!20}{\parbox{\dimexpr\linewidth-13\fboxsep}{Beyond the Visible (BtV) Detection}}}
\label{algo:three}
\KwData{Radar point cloud $R_t$ with measurements $(r_k, \theta_k, \phi_k, d_k, s_k)$ at time $t$}
\KwResult{Metallic detection map $M_{\text{radar}}$}

\textbf{Radar Triggering}:\;
\eIf{RGB-D detects: Wall $||$ occlusion (furniture)}{
    Activate radar\;
}{
    \Return \; (Exit if no occlusion)\;
}

\textbf{SNR Filtering}:\;
\For{each radar measurement in $R_t$}{
    \If{$s_k \geq s_{\text{th}}$}{
        Retain measurement $r_k$\;
    }
}
Let $R_t^{\text{metal}}$ be retained measurements (Eq.~\ref{eq:13})\;

\textbf{Global Transformation}:\;
\For{each measurement in $R_t^{\text{metal}}$}{
    Transform to global frame (Eq.~\ref{eq:14})\;
}

\textbf{Noise Removal}:\;
Outlier removal get $M_{\text{radar}}$ (Eq.~\ref{eq:15})\;

\Return $M_{\text{radar}}$ \tcp*{Radar Point Cloud Map}
\end{algorithm}

\begin{figure*}[]
    \centering
    \includegraphics[width=0.99\textwidth]{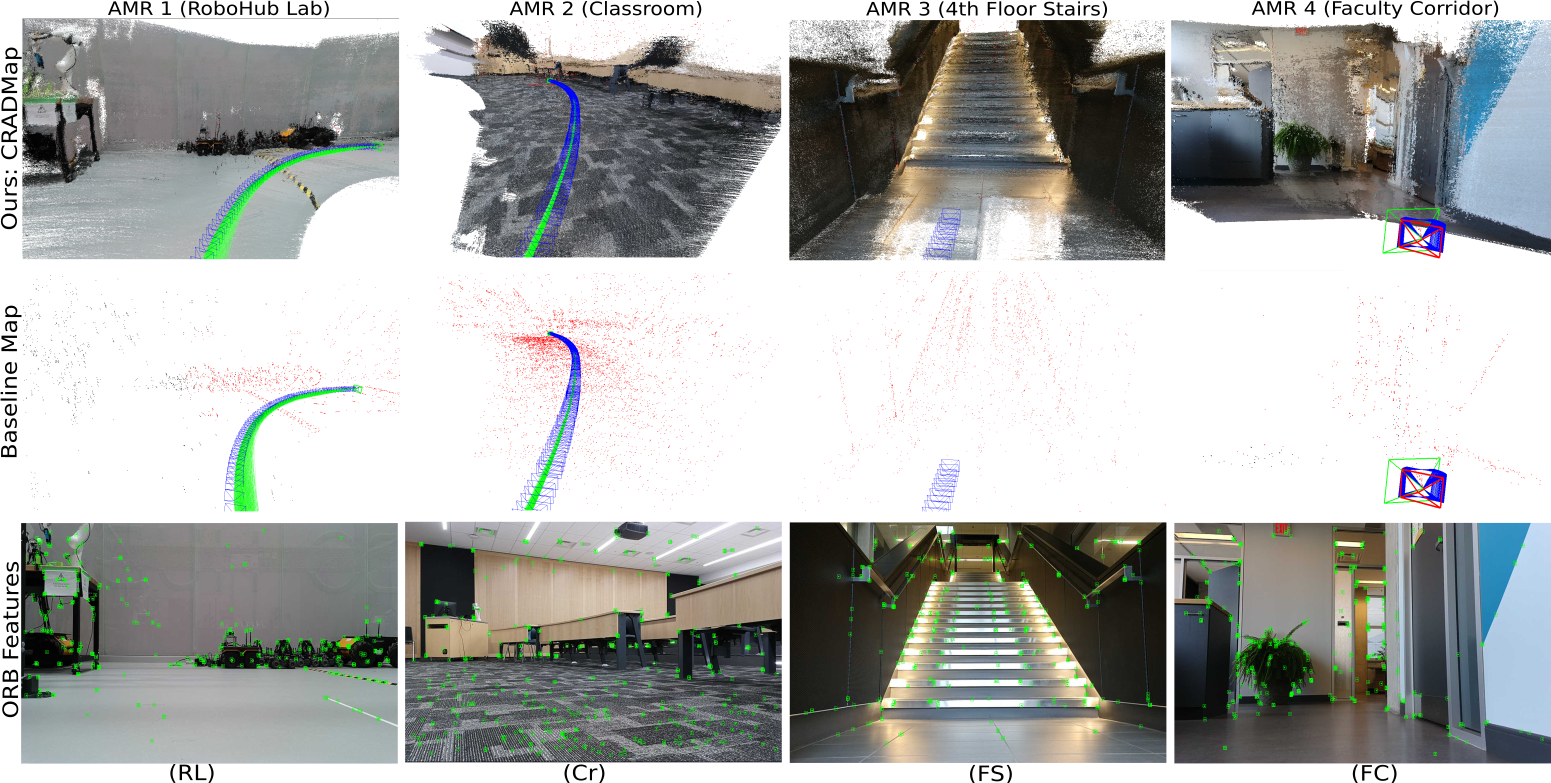}
    \caption{Qualitative comparison of our dense volumetric mapping approach (top row) with the baseline ORB-SLAM3 \cite{campos2021orb} sparse map (middle row). The bottom row shows the field of view and ORB feature detection by the SLAM frontend. The dataset was captured live on different floors of the UW Engineering (UW-E7) building for evaluation.}
    \label{fig:3}
\end{figure*}

\section{Experiments} \label{sec:IV}

\subsection{Evaluation Metrics and Dataset}

\begin{figure*}[]
    \centering
    \includegraphics[width=1.0\textwidth]{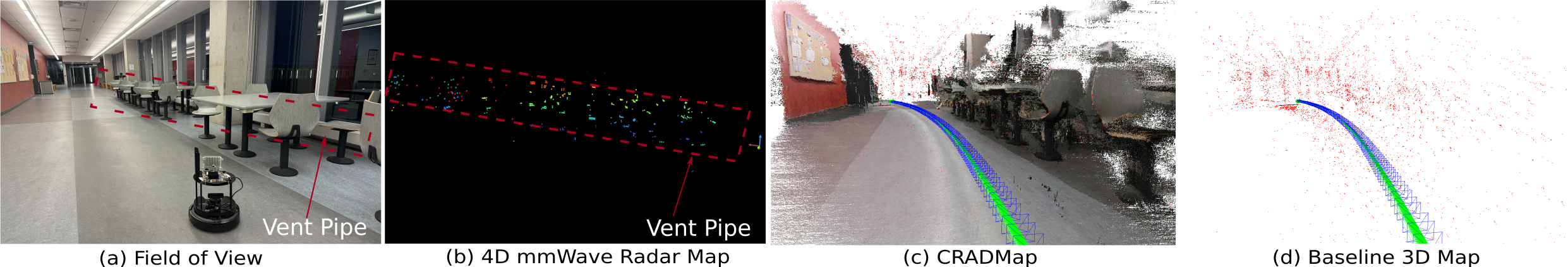}
    \caption{Cluttered indoor setting with furniture and vent pipe present horizontal with the floor. Only 4D mmWave radar detects successfully in (b).}
    \label{fig:4}
\end{figure*}

\begin{figure*}
    \centering
    \includegraphics[width=1.0\textwidth]{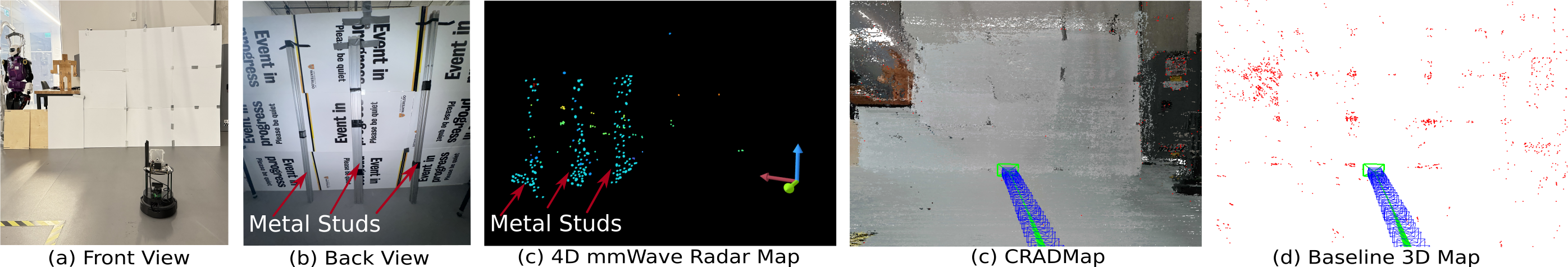}
    \caption{In a completely blocked view, 4D mmWave radar sensing shows a point cloud map of three metal studs behind the handcrafted wall in (c).}
    \label{fig:5}
    \vspace*{-0.1cm} 
\end{figure*}

We evaluate our system on the indoor settings (1) live experiments from the University of Waterloo (UW) E7 building and (2) TUM RGB-D benchmark \cite{sturm12iros}. In UW-E7 we captured 4 scenes as shown in Fig. \ref{fig:3} using four AMRs. All experiments are conducted on hardware Intel i7‑1355U, 32GB RAM, Graphics (RPL-U), and software Ubuntu 22.04, ROS2 Humble, with ORBSLAM3 as our state-of-the-art (SOTA) baseline. Volumetric mapping quality, CRADMap, and network performance are assessed on the UW-E7 dataset in Open3D \cite{Zhou2018} on campus Wifi eduroam \cite{uwaterloo_eduroam} and 5G network  \cite{rogers_website}. 4D mmWave radar detection are evaluate by doing two complex experiments in UW-E7 shown in Fig. \ref{fig:4} and Fig. \ref{fig:5}, compared with Vicon Motion-Capture. Lastly, pose estimation metrics performed on a public dataset.

\subsection{CRADMap Evaluation}
We generated four distributed volumetric maps on live UW-E7 experiments (Robohub Lab, Classroom, Stairs, and Faculty Corridor) using our CRADMap approach. To evaluate mapping, the qualitative comparison is shown in Fig. \ref{fig:3} also we use Open3D for quantitative evaluation to compute coverage (the percentage of the environment reconstructed via a voxel grid over the explored area) and density (points per cubic meter). First, the baseline ORBSLAM3 (SOTA) and our CRADMap are generated to ensure both are in the same coordinate frame, making our relative metrics robust. For example, in the Robohub Lab, glass surfaces cause SOTA to achieve only about 26.91\% coverage, while our method reconstructs approximately 78.93\% of the environment with significantly higher point density as shown in Table \ref{table:2}. Similarly, in the Classroom, Stairs, and Faculty Corridor, our method consistently yields higher coverage and denser reconstructions than the baseline. These relative improvements in coverage 78\% to 85\% and $4.8–5.5\times$ more density provide compelling evidence of our system’s enhanced mapping capabilities, while maintaining real-time performance in Table \ref{tab:system_comparison} and offloads computation load from AMRs.

\begin{table}[h]
\caption{CRADMap Volumetric Mapping Quality Quantitative Analysis on UW-E7 Experiments}
\label{tab:dense_mapping}
\centering
\begin{tabular}{cccccc}
\hline
\rowcolor{gray!20}
 & \multicolumn{2}{c}{\textbf{Coverage (\%)}} & \multicolumn{3}{c}{\textbf{Density (pts/m$^3$)}} \\ \cline{2-3} \cline{5-6}
 \rowcolor{gray!20}
\textbf{Scene} & \textbf{Baseline} & \textbf{Ours} && \textbf{Baseline} & \textbf{Ours} \\ \hline
Robohub Lab & 26.91 & 78.93& & 161 & 950 \\
Classroom & 30.26 & 82.82 && 147 & 896 \\
Stairs & 26.69 & 85.17 && 123 & 782 \\
Faculty Corridor & 24.54 & 84.34 && 182 & 1123 \\ \hline
\label{table:2}
\end{tabular}
\end{table}

\begin{table}[h]
\caption{System-Level Benchmark Comparison on CRADMap (Server Execution vs. AMR Nodes)}
\label{tab:system_comparison}
\centering
\scriptsize
\setlength{\tabcolsep}{4pt}
\begin{tabular}{@{}llccc@{}}
\toprule
\rowcolor{gray!20}
\textbf{Category} & \textbf{Metric} & \textbf{SOTA} & \textbf{CRADMap} & \textbf{AMRs} \\ 
\rowcolor{gray!20}
 & \textbf{Usage} & \textbf{(Server)} & \textbf{(Server)} & \textbf{(Nodes)} \\ 
\midrule

\textbf{Accuracy} 
    & Coverage (\%) & 24.5--30.2 & 78.9--85.1 & -- \\ 
    & Density (pts/m\textsuperscript{3}) & 123--182 & 782--1123 & -- \\ 
\hline

\textbf{Efficiency} 
    & CPU (\%) & 62.4 & 78.9 & 22.7 \\ 
    & (cores) & (4 cores) & (8 cores) & (2 cores) \\
    & GPU (\%) & $<$1 & 52.3±2.5 & 15.4±1.2 \\ 
    & RAM (MB) & 1850±120 & 4250±150 & 410±30 \\ 
\hline

\textbf{Performance} 
    & Frame Process Time (ms) & 22.4±1.8 & 28.6±2.3 & 8.2±0.9 \\ 
    & Update Rate (Hz) & 3.5 (Wifi) & 7.1 (5G) & -- \\ 
    & Upload Rate (MB/s) & 14 & 14 & 14 \\ 
\bottomrule

\multicolumn{5}{@{}p{0.95\columnwidth}}{\footnotesize \textit{Note:} Server as computational node: 14-core i7-11800H + RPL-U GPU; AMRs as publishing ROS2 topics/nodes only i.e. RGB-D, Radar, IMU; Measurements averaged over 10 runs.}
\end{tabular}
\end{table}

\subsection{Network Performance Evaluation}

Evaluation is perform on live UW-E7 dataset in a multi-robot setup, where each AMR generates visual data at 14 MB/s (112 Mbps) from its RGB (15 Hz) and depth (10Hz) streams. The WiFi network at the UW employs IEEE 802.1X for authentication and both dual-band 2.4 GHz/5 GHz frequency bands. For the evaluation, the 5 GHz and WiFi band are utilized, a single AMR achieves an effective maximum upload speed of 90 Mbps, resulting in an update frequency of approximately 8.9 Hz; however, with 4 AMRs the total load reaches 56MB/s (448 Mbps), and the effective per AMR bandwidth drops to about 22.5 Mbps, which get reduced to around 3.5 Hz using 4 AMRs together. In contrast, 5G (3.5 GHz band, 110 to 120 Mbps upload) maintains a stable latency of 24 ms; with 1 AMR, the update frequency is about 10.4 Hz, and with 4 AMRs, it drops to 7.1 Hz due to shared bandwidth as shown in Table \ref{table:3}. These calculations, based on the formula Map Update Frequency = (Effective Upload Speed per AMR)/(Data per Update), clearly show that 5G’s higher available bandwidth and stable latency yield significantly better and more predictable performance.

\begin{table}[htbp]
\caption{Network Performance Quantitative  Comparison on UW-E7 Experiments}
\label{table:3}
\centering
\begin{tabular}{cccccc}
\hline
\rowcolor{gray!20}
 & \multicolumn{2}{c}{\textbf{Data Rate (MB/s)}} & \multicolumn{3}{c}{\textbf{Performance}} \\ \cline{2-3} \cline{5-6}
 \rowcolor{gray!20}
\textbf{Network} & \textbf{AMRs} & \textbf{Total} && \textbf{Latency (ms)} & \textbf{Map (Hz)} \\ \hline
WiFi & 1 & 14 && 8--33 (variable) & 8.9 \\
     & 4 & 56 && 11--34 (variable) & 3.5 \\ \hline
5G   & 1 & 14 && 24 ± 2 & 10.4 \\
     & 4 & 56 && 24 ± 4 & 7.1 \\ \hline
\end{tabular}
\end{table}

\subsection{Beyond the Visible (BtV)-Radar Perception}\label{IVD}
Behind-the-wall object detection enhance perception beyond visual occlusions \cite{lee2020imaging}. To leverage this capability, we automate camera-driven situational awareness to dynamically activate radar when a wall or occlusion region is detected using our Algorithm~\ref{algo:three}. We evaluate 4D mmWave radar capabilities using a 4‑chip cascaded imaging radar operating at 77 GHz, which produces high-density point clouds with an angular resolution of 1.4 degrees in both azimuth and elevation. In the (1) scenario, an AMR maps a cluttered indoor aisle at UW-E7, as shown in Fig. \ref{fig:4}, where the presence of furniture caused the CRADMap in Fig. \ref{fig:4}(c) and baseline SOTA in Fig. \ref{fig:4}(d) to miss detecting a horizontal metal ventilation pipe. Missing such structural components, particularly for AMRs navigating environments, as undetected obstacles may cause collisions. The radar achieved a 92\% ±3 detection rate for the pipe despite some residual noise. Another (2) scenario from Fig. \ref{fig:5}(a,b), 3 cm thick, solid opaque wall made of plastic boards are used to completely block the view. Visibility beyond barriers is critical in SMART factories and inspections. Ground truth are establish via manual annotation placing marker points on the metal studs behind the wall and vent pipe. After comparison with Vicon Motion-Capture, the radar accurately detected metal studs, achieving a 84\% ±5 detection. Radar solve the challenge where traditional visual nodes i.e. camera and lidar fail.

\subsection{Pose Estimation Evaluation}
We evaluate on four TUM RGB-D sequences (fr1/plant, fr1/teddy, fr2/coke, fr2/dishes) to validate pose estimation consistency, as shown in Table~\ref{table:4}. ATE and RMSE were computed by averaging results over five independent runs with randomized initializations, following standard evaluation practices in SLAM literature. Our method slightly improves ATE and RMSE compared to the baseline, as the COVINS backend enables global pose-graph optimization, aggregating loop closures to reduce drift across the full trajectory. Dense depth priors from volumetric fusion further strengthen bundle adjustment in texture-less or reflective regions (e.g., glass walls), where sparse feature methods typically struggle. These challenges are mitigated through the global optimization and loop closures in the COVINS backend.

\begin{table}[htbp]
\caption{Trajectory Accuracy on TUM RGB-D Sequences}
\label{table:4}
\centering
\begin{tabular}{cccc}
\hline
\rowcolor{gray!20}
\textbf{Sequence} & \textbf{Metric} & \textbf{SOTA (m) ± $\sigma$} & \textbf{Ours (m) ± $\sigma$} \\ \hline
fr1/plant  & ATE  & 0.1178 ± 0.0049 & 0.1043 ± 0.0045 \\
                    & RMSE & 0.1323 ± 0.0052 & 0.1187 ± 0.0050 \\ \hline
fr1/teddy  & ATE  & 0.1432 ± 0.0051 & 0.1275 ± 0.0048 \\
                    & RMSE & 0.1654 ± 0.0060 & 0.1482 ± 0.0057 \\ \hline
fr1/coke   & ATE  & 0.1012 ± 0.0043 & 0.0905 ± 0.0039 \\
                    & RMSE & 0.1135 ± 0.0048 & 0.1116 ± 0.0044 \\ \hline
fr1/dishes & ATE  & 0.1476 ± 0.0059 & 0.1398 ± 0.0054 \\
                    & RMSE & 0.1627 ± 0.0061 & 0.1635 ± 0.0056 \\ \hline
\end{tabular}
\end{table}

\section{CONCLUSIONS} \label{sec:V}

This work presented CRADMap, a novel distributed volumetric mapping framework that unifies sparse and dense SLAM methods for multi-robot systems, enabling the detection of Beyond the Visible (BtV) hidden structures. By extending feature-based SLAM with volumetric keyframes and leveraging a 5G-connected backend for global pose-graph optimization, CRADMap delivers globally consistent 3D reconstructions while offloading intensive computation from resource-constrained AMRs. The use of 5G ensures high-bandwidth, enabling real-time performance and a $2\times$ faster map update rate. Experimental results demonstrate substantial improvements, achieving 75\% to 85\% environmental coverage and $4.8\times$ to $5.5\times$ higher point density compared to the SOTA. The integration of a standalone 4D mmWave radar module enables the detection of occluded metallic structures, with detection rates of 84\% to 92\% in cluttered environments where camera-only systems fail. Future work will explore the scalability of mmWave 5G and extend testing to outdoor, including comparisons with other dense SLAM methods.







\bibliography{reference}
\bibliographystyle{ieeetr}

\end{document}